\pdfoutput=1

\documentclass[11pt]{article}

\usepackage[]{acl}

\usepackage{times}
\usepackage{latexsym}

\usepackage[T1]{fontenc}

\usepackage[utf8]{inputenc}

\usepackage{microtype}

\usepackage{graphicx}
\usepackage{multicol}
\usepackage{multirow}
\usepackage{amsmath}

\title{Using Context-to-Vector with Graph Retrofitting to Improve Word Embeddings}

\author{Jiangbin Zheng\textsuperscript{\dag}, Yile Wang\textsuperscript{\dag}, Ge Wang, Jun Xia, \\
    \textbf{Yufei Huang, Guojiang Zhao, Yue Zhang, Stan Z. Li\textsuperscript{*}} \\
    School of Engineering, Westlake University \\
    Institute of Advanced Technology, Westlake Institute for Advanced Study \\
  \texttt{\{zhengjiangbin,wangyile,wangge,xiajun,huangyufei,} \\
  \texttt{zhaoguojiang,zhangyue,Stan.ZQ.Li\}@westlake.edu.cn}
}

\begin{document}
\maketitle
\begin{abstract}
Although contextualized embeddings generated from large-scale pre-trained models perform well in many tasks, traditional static embeddings (e.g., Skip-gram, Word2Vec) still play an important role in low-resource and lightweight settings due to their low computational cost, ease of deployment, and stability. In this paper, we aim to improve word embeddings by 1) incorporating more contextual information from existing pre-trained models into the Skip-gram framework, which we call \textit{Context-to-Vec}; 2) proposing a post-processing retrofitting method for static embeddings independent of training by employing priori synonym knowledge and weighted vector distribution. Through extrinsic and intrinsic tasks, our methods are well proven to outperform the baselines by a large margin.
\end{abstract}

\let\thefootnote\relax\footnotetext{\dag \  Co-first author.}
\let\thefootnote\relax\footnotetext{* Corresponding author.}

\section{Introduction}
Contextualized embeddings such as BERT~\citep{devlin2018bert} and RoBERTa~\citep{2019RoBERTa} have become the default architectures for most downstream NLP tasks. However, they are computationally expensive, resource-demanding, hence environmentally unfriendly. Compared with contextualized embeddings, static embeddings like Skip-gram~\citep{mikolov2013efficient} and GloVe~\citep{pennington2014glove} are lighter and less computationally expensive. Furthermore, they can even perform without significant performance loss for context-independent tasks like lexical-semantic tasks (e.g., word analogy), or some tasks with plentiful labeled data and simple language~\citep{arora2020contextual}.

\begin{figure}[htb]
	\centering
	\includegraphics[width=0.7\columnwidth]{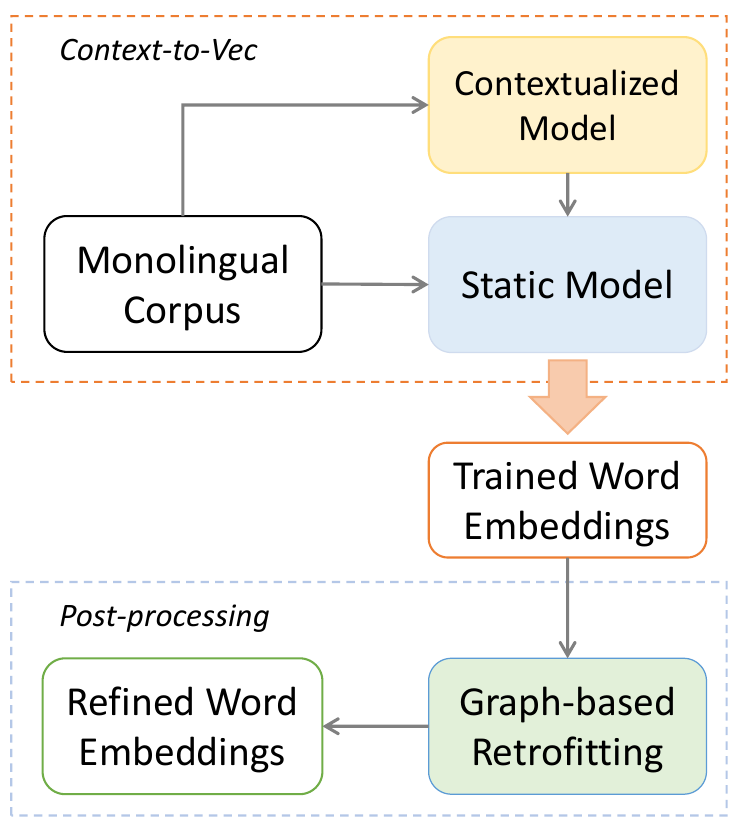}
	\caption{The overall training pipeline of our proposed word embeddings training and post-processing methods. In the \text{Context-to-Vec} phase, static word embeddings are trained using contextualized embeddings based on a monolingual corpus. While in the post-processing phase, external knowledge is introduced to fine-tune the word vectors based on the graph topology.}
	\label{fig_1}
\end{figure}

Recent work has attempted to enhance static word embedding while maintaining the benefits of both contextualized embedding and static embedding. Among these efforts, one category is the direct conversion of contextualized embeddings to static embeddings~\citep{bommasani2020interpreting}. The other category of enhancement is to make use of contextualized embeddings for static embeddings~\citep{melamud2016context2vec}. The latter category is a newer paradigm, which we call \textit{Context-to-Vec}. This paradigm not only alleviates the word sense ambiguities from static embedding, but also fuses more syntactic and semantic information in the context within a fixed window.

For the \textit{Context-to-Vec} paradigm, an association between contextualized word vectors and static word vectors is essentially required. In this case, the contextualized signal serves as a source of information enhancement for the static embeddings~\citep{vashishth2018incorporating}. However, the existing efforts only consider the contextualized embeddings of center words as the source, which is actually incomplete since the contextualized features for the context words of the center words are ignored.

In addition, benefiting from the invariance and stability of already trained static embeddings, post-processing for retrofitting word vectors is also an effective paradigm for improving static embeddings. For example, one solution is an unsupervised approach that performs a singular value decomposition to reassign feature weights~\citep{artetxe2018uncovering}, but this does not utilize more external knowledge and lacks interpretation. Poor initial spatial distribution of word embeddings obtained from training may lead to worse results. Another common solution is to use a synonym lexicon~\citep{faruqui2014retrofitting}, which exploits external prior knowledge with more interpretability but does not take into account the extent of spatial distance in the context.

In this work, we unify the two paradigms above within a model to enhance static embeddings. On the one hand, we follow the \textit{Context-to-Vec} paradigm in using contextualized representations of center words and their context words as references for static embeddings. On the other hand, we propose a graph-based semi-supervised post-processing method by using a synonym lexicon as prior knowledge, which can leverage proximal word clustering signals and incorporate distribution probabilities. The overall training pipeline is shown in Fig.\ref{fig_1}. The pipeline is divided into two separate phases, where the first phase follows the \textit{Context-to-Vec} paradigm by distilling contextualized information into static embeddings, while the second phase fine-tunes the word embeddings based on graph topology. To validate our proposed methods, we evaluate several intrinsic and extrinsic tasks on public benchmarks. The experimental results demonstrate that our models significantly outperform traditional word embeddings and other distilled word vectors in word similarity, word analogy, and word concept categorization tasks. Besides, our models moderately outperform baselines in all downstream clustering tasks.

To our knowledge, we are the first to train static word vectors by using more contextual knowledge in both training and post-processing phases. The code and trained embeddings are made available at \url{https://github.com/binbinjiang/Context2Vector}.

\section{Related Work}
\textbf{Word Embeddings}. For traditional static word embeddings, Skip-gram and CBOW are two models based on distributed word-context pairs~\citep{mikolov2013efficient}. The former uses center words to predict contextual words, while the latter uses contextual words to predict central words. GloVe is a log-bilinear regression model which leverages global co-occurrence statistics of corpus~\citep{pennington2014glove}; FASTTEXT takes into account subword information by incorporating character n-grams into the Skip-gram model~\citep{bojanowski2017enriching}. While contextualized word embeddings~\citep{peters2018deep,devlin2018bert} have been widely used in modern NLP. These embeddings are actually generated using language models such as LSTM and Transformer~\citep{vaswani2017attention} instead of a lookup table. This paradigm can generally integrate useful sentential information into word representations.

\textbf{Context-to-Vec}. The fusion of contextualized and static embeddings is a newly emerged paradigm in recent years. For instance, \citet{vashishth2018incorporating} propose SynGCN using GCN to calculate context word embeddings based on syntax structures; \citet{bommasani2020interpreting} introduce a static version of BERT embeddings to represent static embeddings; \citet{wang2021improving} enhance the Skip-gram model by distilling contextual information from BERT. Our work also follows this paradigm but introduce more context constraints.

\textbf{Post-processing Embeddings}. Post-processing has been used for improving trained word embeddings. Typically, \citet{faruqui2014retrofitting} use synonym lexicons to constrain the semantic range; \citet{artetxe2018uncovering} propose a method based on eigenvalue singular decomposition. Similar to these techniques, our post-processing method is easy for deployment and can be applied to any static embeddings. The difference is that we not only take advantage of the additional knowledge, but also consider the distance weights of the word vectors, overcoming the limitations of existing methods with better interpretability.

\section{Proposed Methods}
\begin{figure*}
	\centering
	\includegraphics[width=2.1\columnwidth]{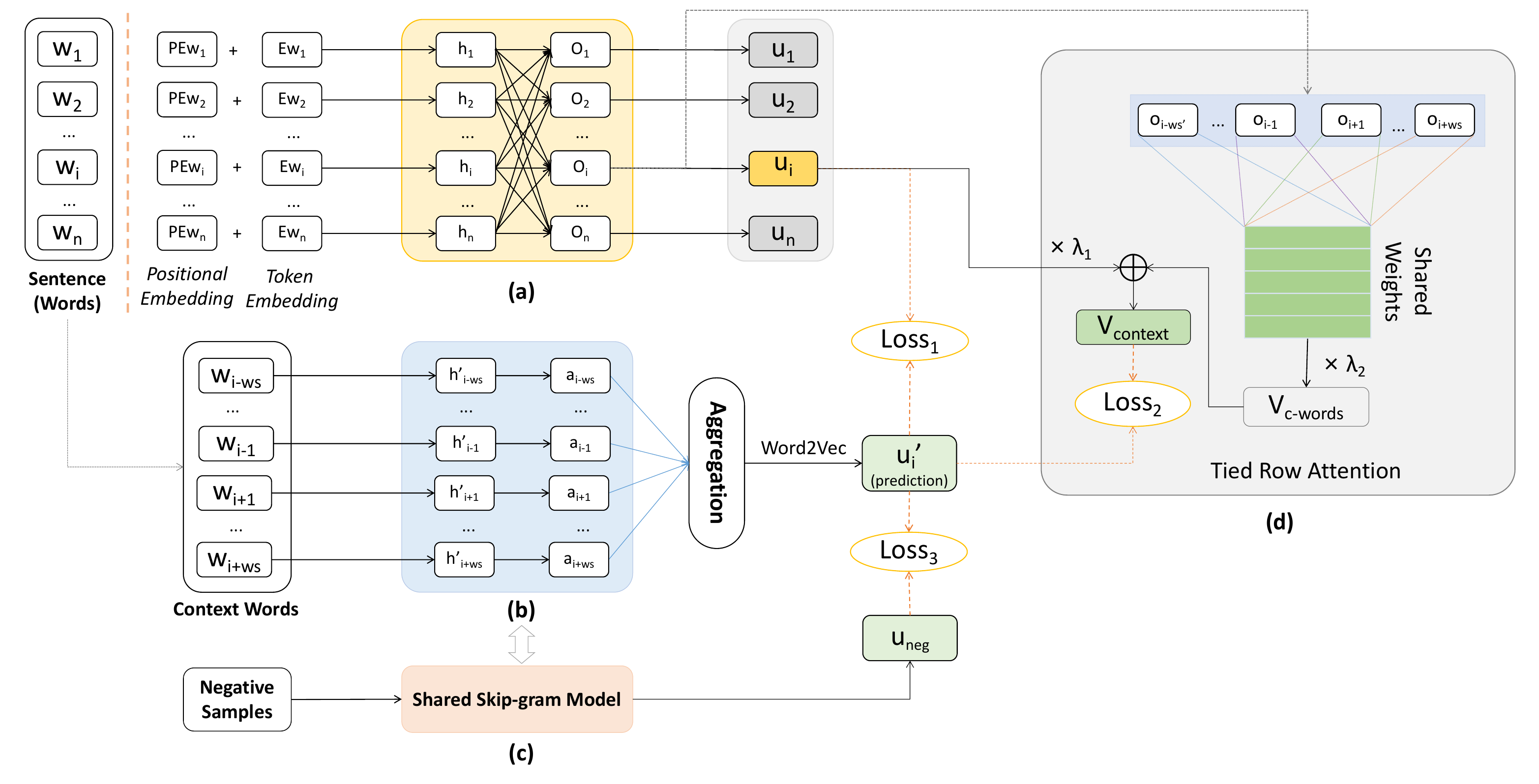}
	\caption{Main framework of our model. (a) \textbf{Contextual Embedding Module} generally consists of a pre-trained language model (BERT-like models) that provides the main enhancement information for static embeddings; (b) \textbf{Static Embedding Module} is the core component for training word embeddings from scratch and obtaining distilled contextualized information; (c) \textbf{Negative Sample Module} collects negative samples randomly and constructs contrast loss to improve the robustness and generalization; (d) \textbf{Tied Contextualized Attention Module} is to capture the contextualized embeddings for the context words as supplementary information.}
	\label{fig_2}
\end{figure*}

\subsection{Embedding Representations}
As shown in Fig.\ref{fig_2}, our proposed framework consists of four basic components. Formally, given a sentence $s=\{w_1, w_2, ... , w_n\} (w_i \in D)$, our objective is to model the relationship between the center word $w_i$ and its context words $\{ w_{i-w_s}, ... , w_{i-1}, w_{i+1}, ... , w_{i+w_s}\}$.

\textbf{Contextualized Embedding Module}. To incorporate contextualized information, an embedding $u_i$ of the center word $w_i$ needs to be generated from a pre-trained language model (Fig.\ref{fig_2}(a)). Taking the BERT model as an example, the center word $w_i$ is first transformed into a latent vector $h_i$, then $h_i$ is fed to a bidirectional Transformer for self-attention interaction. Finally, the output representation $o_i \in R^d$ is linearly mapped to $u_i\in R^{d_{emb}}$ through a linear layer as:
\begin{equation}
u_i = W_o {\rm Linear} ({\rm SA} (h_i)) = W_o o_i,
\label{eq1}
\end{equation}
where $W_o \in R^{d_{emb}\times d}$ denotes model parameters, ${\rm Linear(*)}$ denotes a linear mapping layer, and ${\rm SA(*)}$ denotes self-attention. In practice, the size of $o_i$ is $d=768$, and the size of $u_i$ is $d_{emb}=300$. The $h_i$ here is a sum of the \textit{Token Embedding} $E_{w_i}$ and the \textit{Positional Embedding} $PE_{w_i}$ as:
\begin{equation}
h_i = E_{w_i} + PE_{w_i}.
\end{equation}

\textbf{Static Embedding Module}. The Skip-gram model (Fig.\ref{fig_2}(b)) is used as the static embedding module. Our method does not directly fit the Skip-gram model by replacing an embedding table, although the original Skip-gram uses an embedding table of center words as the final embedding. Instead, to make the context words predictable and to enable negative sampling from the vocabulary, contextualized representations are used for the center words, while an embedding table of the context words is used for the output static embedding.

\subsection{Heuristic Semantic Equivalence}
As mentioned above, a key issue for the \textit{Context-to-Vec} paradigm is to bridge the gap between contextualized and static word vectors. To this end, a main intuition is to find key equivalent semantic connections between contextualized vectors and static vectors. We take the following heuristics:

\textbf{Heuristic 1}: \textit{For a given sentence, the contextualized embedding representation of a center word can be semantically equivalent to the static embedding of the center word in the same context}.

According to \textbf{Heuristic 1}, in order to model the center word $w_i$ and its context words $w_{i+j}$ (note here that the illegal data that indexes less than 0 or greater than the maximum length are ignored), a primary training target is to maximize the probability of the context words $w_{i+j}(|j| \in [1, w_s])$ in the Skip-gram model:
\begin{equation}
p(w_{i+j}|w_i)=\frac{{\rm exp}({{u'}_{i+j}}^{T} u_i)}{\sum_{w_k \in D}{\rm exp}({{u'}_k}^T u_i)},
\end{equation}
where $u_i$ is the contextualized representation of the center word, and ${u'}_k$ is the static embedding from a center word $w_k$ that is generated by a static embedding table with size $d'=300$.

For \textbf{Heuristic 1}, the contextualized word embedding of any center word is essentially used as reference for corresponding static word embedding. Such a source for information enhancement implicitly contains the context of the contextualized embedding, but explicitly ignores the contextual information which is easily accessible. Hence, the proposed:

\textbf{Heuristic 2}: \textit{Inspired by the idea of Skip-gram-like modeling, the contextualized embedding representation for the context words of a center word can be also semantically equivalent to represent the static embedding of the center word}.

To model this semantic relationship, we introduce a \textbf{Tied Contextualized Attention} module (Fig.\ref{fig_2}(d)) for explicitly attending contextual signals, which complements \textbf{Heuristic 1} by incorporating more linguistic knowledge into the static embedding. In particular, assume that the center word $w_i$ in the contextualized embedding module corresponds to the contextual vocabulary notated as $\{w_{i-w_s'},... , w_{i-1}, w_{i+1}... ,w_{i+w_s'}\}$, then the output contextual attention vector can be computed as:
\begin{equation}
\begin{aligned}
V_{context} &=\lambda_1 V_{center} + \lambda_2 V_{c-words} \\
& = \lambda_1 o_i^T W_1 +  \lambda_2 \tau( U_{1 \leq |k| \leq i+w_s’} \phi(o_k^T W_2)) \\
&=\lambda_1 o_i^T W_1+ \lambda_2 \frac{\phi(\sum_{1 \leq |k| \leq i+w_s’} o_k^T)} {2 w_s’} W_2,
\end{aligned}
\label{eq4}
\end{equation}
where $V_{center}$ denotes the embedding representation of the center word, which is a residual connection here. And $V_{c-words}$ denotes the embedding representations of corresponding context words. $\phi$ is an optional nonlinear function, $U(*)$ is a merge operation, and $\tau$ is an average pooling operation. $W_1 \in R^{d \times d_{emb}}$ and $W_2 \in R^{d \times d_{emb}}$ are trainable parameters, in which $W_2$ denotes the weight assignment of each context vector.

Since each ${o}_k$ has similar linguistic properties, the weight $W_2$ can be shared, and we name this module \textbf{Tied Contextualized Attention} mechanism. Therefore, the weighted average of the linear transformation of all context vectors can be reduced to the weighted linear output of the average of all vectors as shown in Eq.\ref{eq4}. This weight-sharing mechanism can help speed up calculations.

In practice, to reduce the complexity, the weight parameter $\lambda_{1}$ and $\lambda_{2}$ are the same; the $u_i$ in Eq.\ref{eq1} can be directly used as $V_{context}$; the value of $w_s’$ is the same as that of $w_s$, e.g., 5.

\subsection{Training Objectives}
The modular design requires our model to satisfy multiple loss constraints simultaneously, allowing static embeddings to introduce as much contextual information as possible. Given a training corpus with $N$ sentences $s_c = \{w_1, w_2,... , w_{n_c} \}(c \in [1,N])$, our loss functions can be described as follows.

\textbf{Semantic Loss}. As illustrated in \textbf{Heuristic 1}, one of our key objectives is to learn the semantic similarity between the contextualized embedding and the static embedding of the center word. To speed up computation, the inner product of the normalized vectors can be used as the loss $L_1$:
\begin{equation}
L_1 = -\sum_{c=1}^{N}\sum_{i=1}^{n_c}({\rm log}\sigma({(\sum_{1 \leq |j| \leq w_s} {u’}_{i+j})}^{T} u_i)],
\end{equation}
where $\sigma$ is the sigmoid function.

\textbf{Contextualized Loss}. As described in \textbf{Heuristic 2}, the contextualized embeddings for the context words of the center word are explicitly introduced to further enhance the static embedding, thus the Contextualized Loss $L_2$ is expressed as:
\begin{equation}
L_2 = -\sum_{c=1}^{N}\sum_{i=1}^{n_c}({\rm log}\sigma(V_{context}^{T}u_i)).
\end{equation}

\textbf{Contrastive Negative Loss}. Negative noisy samples (Fig.\ref{fig_2}(c)) can improve the robustness and effectively avoid the computational bottleneck. This trick is common in NLP. Our Contrastive Negative Loss $L_3$ is calculated as:
\begin{equation}
L_3 = \sum_{c=1}^{N}\sum_{i=1}^{n_c}\sum_{m=1}^{k} E_{w_{{neg}_m}~P(w)}[{\rm log}\sigma({u’}_{{neg}_m}^{T}u_i)],
\end{equation}
where $w_{{neg}_m}$ denotes a negative sample, $k$ is the number of negative samples and $P(w) $is a noise distribution set. 

\textbf{Joint Loss}. The final training objective is a joint loss $L$ for multi-tasks as:
\begin{equation}
L = \eta_{1} L_1+ \eta_{2} L_2+ \eta_{3} L_3,
\end{equation}
where each hyperparameter $\eta_{i}$ denotes a weight.

\begin{figure}
	\centering
	\includegraphics[width=0.8\columnwidth]{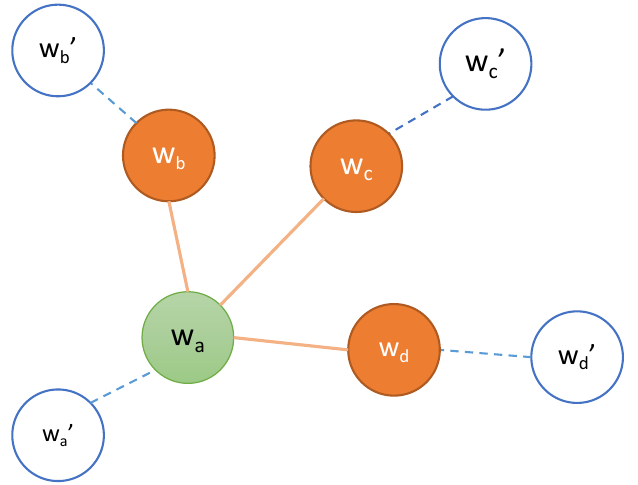}
	\caption{A word graph diagram with edges between related words. The dashed edges indicate the corresponding edge relationships between observed word vectors (white nodes) and inferred word vectors (colored nodes). And the solid edges indicate the relationship between the word (green node) to be refined and its corresponding synonyms (orange nodes).}
	\label{fig_3}
\end{figure}

\subsection{Graph-based Post-retrofitting}
In the post-processing stage, we propose a new semi-supervised retrofitting method for static word embeddings based on graph topology~\citep{xia2022survey,wu2021self,wu2020stc}. This method overcomes the limitations of previously existing work by 1) using a synonym lexicon as priori external knowledge. Since both contextualized embeddings and static embeddings are trained in a self-supervised manner, the word features originate only from within the sequence and no external knowledge is considered; 2) converting the Euclidean distances among words into a probability distribution~\citep{mcinnes2018umap}, which is based on the special attributes that the trained static word vectors are mapped in a latent Euclidean space and remain fixed.

\textbf{Word Graph Representation}. Suppose that $V = \{w_1, ... , w_n\}$ is a vocabulary (i.e., a collection of word types). We represent the semantic relations among words in $V$ as an undirected graph $(V, E)$, with each word type as a vertex and edges $(w_i, w_j) \in E$ as the semantic relations of interest. These relations may vary for different semantic lexicons. Matrix $Q'$ represents the set of trained word vectors for $q'_i \in R^{Dim}$, in which $q'_i$ corresponds to the word vector of each word $w_i$ in $V$. 

Our objective is to learn a set of refined word vectors, denoted as matrix $Q = (q_1, ... , q_n)$, with the columns made close to both their counterparts in $Q'$ and the adjacent vertices according to the probability distribution. A word graph with such edge connectivity is shown in Fig.\ref{fig_3}, which can be interpreted as a Markov random field~\citep{li1994markov}.

\textbf{Retrofitting Objective}. To refine all word vectors close to the observed value $q_i'$ and its neighbors $q_j$ ($(i, j) \in E$), the objective is to minimize:
\begin{equation}
\begin{array}{l}
\begin{aligned}
\Psi(Q)= \sum_{i=1}^{n} (&\alpha_i ||q_i - {q’}_i||^2 \\
+ \beta_i \sum_{(i,j) \in E} &\gamma_{ij}||q_i - q_j||^2),
\end{aligned}
\end{array}
\end{equation}
where $\alpha_i$, $\beta_i$, and $\gamma_{ij}$ control the relative strengths of associations, respectively. Since $\Psi$ is convex in $Q$, we can use an efficient iterative update algorithm. The vectors in $Q$ are initialized to be equal to the vectors in $Q'$. Assuming that $w_i$ has $m$ adjacent edges corresponding to $m$ synonyms, then we take the first-order derivative of $\Psi$ with respect to a $q_i$ vector and equate it to zero, yielding the following online update:
\begin{equation}
q_i = \alpha_i {q’}_i + \beta_i \frac{\sum_{j:(i,j) \in E} \gamma_{ij} q_j}{m}.
\end{equation}

By default, $\alpha_{i}$ and $\beta_{i}$ take the same value 0.5, and $\gamma_{ij}$ can be expressed as:
\begin{equation}
\gamma_{ij}=g(d_{ij}|\sigma,\nu)=C_\nu{(1+\frac{d_{ij}^2}{\sigma\nu})}^{-{(\nu+1)}} \in (0,1],
\end{equation}
in which $\sigma$ is a scale parameter, $\nu$ is a positive real parameter, and $C_\nu$ is the normalization factor of $\nu$ as (the following ${\rm \Gamma(*)}$ denotes the gamma function):
\begin{equation}
C_\nu=2\pi(\frac{{\rm \Gamma}(\frac{\nu+1}{2})}{\sqrt{\nu\pi}{\rm \Gamma}(\frac{\nu}{2})})^2,
\end{equation}
and $d_{ij}$ calculates the sum of Euclidean distances of the feature vectors across all dimensions $Dim$ as:
\begin{equation}
d_{ij}=\sqrt{\sum_{k=0}^{Dim}{(q_{i_k}-q_{j_k})}^2}.
\end{equation}

Through the above process, the distance distribution is first converted into a probability distribution, and then the original word graph is represented as a weighted graph. This retrofitting method is modular and can be applied to any static embeddings.

\begin{table*}
\centering
\scalebox{0.7}{
\begin{tabular}{c|c|ccccccc|cc}

\hline \multirow{2}{*}{\textbf{Types}} & \multirow{2}{*}{\textbf{Models}} & \multicolumn{7}{c|}{\textbf{Word Similarity}} & \multicolumn{2}{c}{\textbf{Analogy}} \\ 

&  & \textbf{WS353} & \textbf{WS353S} & \textbf{WS353R} & \textbf{SimiLex} & \textbf{RW} & \textbf{MEN} & \textbf{RG65} & \textbf{Google} & \textbf{SemEval} \\ \hline

\multirow{6}{*}{\textbf{Static}}	& Skip-gram	& 61.0	& 68.9	& 53.7	& 34.9	& 34.5	& 67.0	& 75.2	& 43.5	& 19.1 \\
	& Skip-gram(context)	& 53.2	& 60.9	& 43.5	& 32.0	& 28.0	& 58.8	& 69.3	& 40.6	& 16.7 \\
	& CBOW	& 62.7	& 70.7	& 53.9	& 38.0	& 30.0	& 68.6	& 72.7	& 58.4	& 18.9 \\
	& GloVe	& 54.2	& 64.3	& 50.2	& 31.6	& 29.9	& 68.3	& 61.8	& 45.3	& 18.7 \\
	& FASTTEXT	& 68.3	& 74.6	& 61.6	& 38.2	& 37.3	& 74.8	& 80.8	& 72.7	& 19.5 \\
	& Deps	& 60.6	& 73.1	& 46.8	& 39.6	& 33.0	& 60.5	& 77.1	& 36.0	& 22.9 \\
	 \hline
	 
\multirow{12}{*}{\textbf{Contextualized}} 	& ELMo$_{token}$	& 54.1	& 69.1	& 39.2	& 41.7	& 42.1	& 57.7	& 69.6	& 39.8	& 19.3 \\
	& GPT2$_{token}$	& 65.5	& 71.5	& 55.7	& 48.4	& 31.6	& 69.8	& 63.2	& 33.1	& 21.3 \\
	& BERT$_{token}$	& 57.8	& 67.3	& 42.5	& 48.9	& 29.5	& 54.8	& 66.1	& 31.7	& 22.0 \\
	& XLNet$_{token}$	& 62.4	& 74.4	& 53.2	& 48.1	& 34.0	& 66.3	& 68.3	& 32.6	& 22.2 \\

    & ELMo$_{word}$	& 45.5	& 62.1	& 32.4	& 40.6	& 34.6	& 57.2	& 60.9	& 36.4	& 22.6 \\
	& GPT2$_{word}$	& 30.7	& 31.4	& 27.6	& 26.4	& 22.5	& 26.2	& 10.6	& 19.9	& 12.5 \\
	& BERT$_{word}$	& 24.0	& 31.0	& 14.1	& 13.4	& 10.8	& 22.0	& 18.5	& 25.2	& 10.1 \\
	& XLNet$_{word}$	& 62.8	& 69.8	& 55.5	& 49.0	& 29.7	& 61.7	& 63.4	& 31.9	& 22.5 \\
	& ELMo$_{avg}$	& 58.3	& 71.3	& 47.4	& 43.6	& 38.4	& 65.5	& 66.8	& 49.1	& 21.2 \\
	& GPT2$_{avg}$	& 64.5	& 72.1	& 59.7	& 46.9	& 29.1	& 68.6	& 80.0	& 37.2	& 21.9 \\
	& BERT$_{avg}$	& 59.4	& 67.0	& 49.9	& 46.8	& 30.8	& 66.3	& 81.2	& 59.4	& 20.8 \\
	& XLNet$_{avg}$	& 64.9	& 72.3	& 58.0	& 47.3	& 27.7	& 64.1	& 69.7	& 30.8	& \underline{23.2} \\ \hline
	
\multirow{5}{*}{\textbf{Context-to-Vec}}	& ContextLSTM	& 63.5	& 66.6	& 57.3	& 39.3	& 23.1	& 66.4	& 72.6	& 60.7	& 20.0 \\
	& SynGCN	& 60.9	& 73.2	& 45.7	& 45.5	& 33.7	& 71.0	& 79.6	& 58.5	& \textbf{23.4} \\
	& BERT+Skip-gram	& 72.8 & 75.3	& 66.7	& 49.4	& 42.3	& 76.2	& 78.6	& 75.8	& 20.2 \\
    &	Ours(preliminary)	& \underline{76.9}	& \underline{76.7}	& \underline{68.3}	& \underline{54.9}	& \underline{43.5}	& \underline{76.8}	& \underline{84.3}	& \underline{75.6}	& 20.3 \\
	& Ours(+post-process)	& \textbf{78.9} & \textbf{77.0}	& \textbf{70.1}	& \textbf{55.2}	& \textbf{44.0}	& \textbf{77.9}	& \textbf{85.1}	& \textbf{76.3}	& 21.4  \\	
\hline
\end{tabular}
}
\caption{\label{table_1} Results on word similarity and analogy tasks. \textit{Ours(preliminary)}: without post-processing; \textit{Ours (+post-process)}: with post-processing. The best results are bolded, and the second-best underlined.}
\end{table*}

\section{Experiments}

We use Wikipedia to train static embeddings. The cleaned corpus has about 57 million sentences and 1.1 billion words. The total number of vocabularies is 150k. Sentences between 10 and 40 in length were selected during training.

\subsection{Evaluation Benchmarks}
We conduct both intrinsic and extrinsic evaluations.

\textbf{Intrinsic Tasks}. We conduct \textbf{word similarity} tasks on the WordSim-353~\citep{finkelstein2001placing}, SimLex-999~\citep{kiela2015specializing}, Rare Word (RW)~\citep{luong2013better}, MEN-3K~\citep{bruni2012distributional}, and RG-65~\citep{rubenstein1965contextual} datasets, computing the Spearman’s rank correlation between the word similarity and human judgments. For \textbf{word analogy} task, we compare the analogy prediction accuracy on the Google~\citep{mikolov2013efficient} dataset. The Spearman’s rank correlation between relation similarity and human judgments is compared on the SemEval-2012~\citep{jurgens2012semeval} dataset. \textbf{Word concept categorization} tasks involves grouping nominal concepts into natural categories. We evaluate on AP~\citep{almuhareb2006attributes}, Battig~\citep{baroni2010distributional} and ESSLI~\citep{baroni2008lexical} datasets. Cluster purity is used as the evaluation metric.

\textbf{Extrinsic Tasks}. The CONLL-2000 shared task~\citep{sang2000introduction} is used for \textbf{chunking} tasks and F1-score is used as the evaluation metric; OntoNotes 4.0~\citep{weischedel2011ontonotes} is used for \textbf{NER} tasks and F1-score is used as the evaluation metric; And the WSJ portion of Penn Treebank~\citep{marcus1993building} is used for \textbf{POS tagging} tasks, and token-level accuracy is used as the evaluation metric. These tasks are reimplemented with the open tool NCRF++~\citep{yang2018ncrf++}.

\subsection{Baselines}
As shown in Table \ref{table_1}, baselines are classified into three categories. For the first category (\textit{Static}), static embeddings come from a lookup table. Note here that \textit{Skip-gram(context)} denotes the results from the context word embeddings. For the second category (\textit{Contextualized}), static embeddings come from contextualized word embedding models (i.e., BERT, ELMo, GPT2, and XLNet) for lexical semantics tasks. The models with \textit{\_token} use the mean pooled subword token embeddings as static embeddings; The models with \textit{\_word} take every single word as a sentence and output its word representation as a static embedding; The models with \textit{\_avg} take the average of output over training corpus. For the last category (\textit{Context-to-Vec}), contextualized information is integrated into Skip-gram embeddings. Among these models, ContextLSTM~\citep{melamud2016context2vec} learns the context embeddings by using single-layer bi-LSTM; SynGCN~\citep{vashishth2018incorporating} uses GCN to calculate context word embeddings based on syntax structures; BERT+Skip-gram~\citep{wang2021improving} enhances the Skip-gram model by adding context syntactic information from BERT, which is our primary baseline.

\begin{table}
\centering
\scalebox{0.65}{
\begin{tabular}{c|cccc|c}
\hline \textbf{Models} & \textbf{AP} & \textbf{Batting} & \textbf{ESSLI(N)} & \textbf{ESSLI(V)} & \textbf{Avg} \\ \hline
Skip-gram	& 63.4	& 42.8	& 75.0	& 62.2	& 60.8 \\
Skip-gram(context)	& 57.4	& 41.6	& 72.5	& 66.6	& 59.5 \\
CBOW	& 63.2	& 43.3	& 75.0	& 64.4	& 61.4 \\
Glove	& 58.0	& 41.3	& 72.5	& 60.0	& 58.0 \\
FASTTEXT	& 63.4	& 44.4	& 75.0	& 62.2	& 61.2 \\
Deps	& 61.8	& 41.7	& 77.5	& 68.8	& 62.4 \\
BERT$_{avg}$	& 55.7	& 34.7	& 70.0	& 64.0	& 56.1 \\
SynGCN	& 63.4	& 42.8	& 82.5	& 62.2	& 62.7 \\
BERT+Skip-gram	& 64.1	& 43.8	& 77.5	& 66.6	& 63.0 \\
Ours(preliminary)	& \underline{65.7}	& \underline{44.0}	& \underline{85.0}	& \underline{70.4}	& \underline{66.3} \\
Ours(+post-process)	& \textbf{66.4}	& \textbf{44.2}	& \textbf{87.5}	& \textbf{74.1}	& \textbf{68.1} \\
\hline
\end{tabular}
}
\caption{\label{table_2} Results on word concept categorization tasks. The best results are bolded, and the second-best underlined.}
\end{table}

\begin{table}
\centering
\scalebox{0.8}{
\begin{tabular}{c|ccc|c}
\hline \textbf{Models} & \textbf{CHUNK} & \textbf{NER} & \textbf{POS} & \textbf{Avg} \\ \hline
Skip-gram 	& 88.07 	&  83.90	&  95.12	&  89.03 \\
GloVe  	& 89.87 	&  89.13	&  96.52	& 91.84    \\
BERT$_{avg}$ 	&  90.96	&  84.51	&  96.80	&  90.76  \\
SynGCN 	& \underline{91.23} 	&  88.75	&  96.71	&  82.23 \\
BERT+Skip-gram 	&  91.06 	& \underline{88.98} 	& \underline{96.86} 	&  \underline{92.30} \\
Ours 	& \textbf{91.98} 	&  \textbf{89.52}	& \textbf{96.91} 	& \textbf{92.80}    \\
\hline
\end{tabular}
}
\caption{\label{table_3} Results on extrinsic tasks. The best results are bolded, and the second-best underlined.}
\end{table}

\begin{table*}[ht]
	\centering
	\scalebox{0.7}{
	\begin{tabular}{c|ccccccc|c}
	\hline \textbf{Methods} & \textbf{WS353} & \textbf{WS353S} & \textbf{WS353R} & \textbf{SimiLex} & \textbf{RW} & \textbf{MEN} & \textbf{RG65}  & Avg \\ \hline
	w/o retrofitting	& 76.9	& 76.7	& 68.3	& 54.9 & 43.5 & 76.8 & 84.3 & 68.8 \\
	+\citet{faruqui2014retrofitting} & 77.2	& 76.1	& 69.8	& 55.0  & 43.8 & 76.2 & 83.5 & 68.8 \\
	+\citet{artetxe2018uncovering}	& 78.3	& 75.3	& 70.0	& 49.4   & 42.7 & 77.4 & 84.6 & 68.2 \\
	+Ours	& \textbf{78.9}	& \textbf{77.0}	& \textbf{70.4}	& \textbf{55.2}  & \textbf{44.0} & \textbf{77.9} & \textbf{85.1}  & \textbf{69.8}\\
	\hline
	\end{tabular}
	}
	\caption{\label{table_4} Comparison on post-processing schemes.}
\end{table*}

\begin{table*}
\centering
\scalebox{0.68}{
\begin{tabular}{c|c|c}
\hline \textbf{Models} & \textbf{Nearest neighbors of \textit{light}} & \textbf{Nearest neighbors of \textit{while}} \\ \hline
Skip-gram & uv, bioluminescence, fluorescent, glare, sunlight, illumination & whilst, recuperating, pursuing, preparing, attenmpting, fending\\
CBOW	 & stevenson, initimadation, earle, yellowing, row, kizer & whilst, when, still, although, and, but \\	
GloVe	 & excluding, justify, orestes, generation, energy, frieze& both, taking, ',' , up, but, after\\	
FASTTEXT	 & sculpts, baha'i, kinghorn, lick, inputs, minimize  & whilst, still, and, meanwhile, instead, though \\	
SynGCN		 & search, prostejov, preceding, forearms, freewheel, naxos & whilst, time, when, years, months, tenures \\
BERT+SkipGram	& lights, dark, lighter, illumination, glow, illuminating & whilst, whereas, although, conversely, though, meanwhile\\
Ours	& lumière, lumiere, licht, illumination, luminous, lights & whilst, whereas, although, though, despite, albeit \\
\hline
\end{tabular}
}
\caption{\label{table_6} Nearest neighbors of words “\textit{light}” and “\textit{while}”.}
\end{table*}

\begin{figure}[ht]
	\centering
	\includegraphics[width=1\columnwidth]{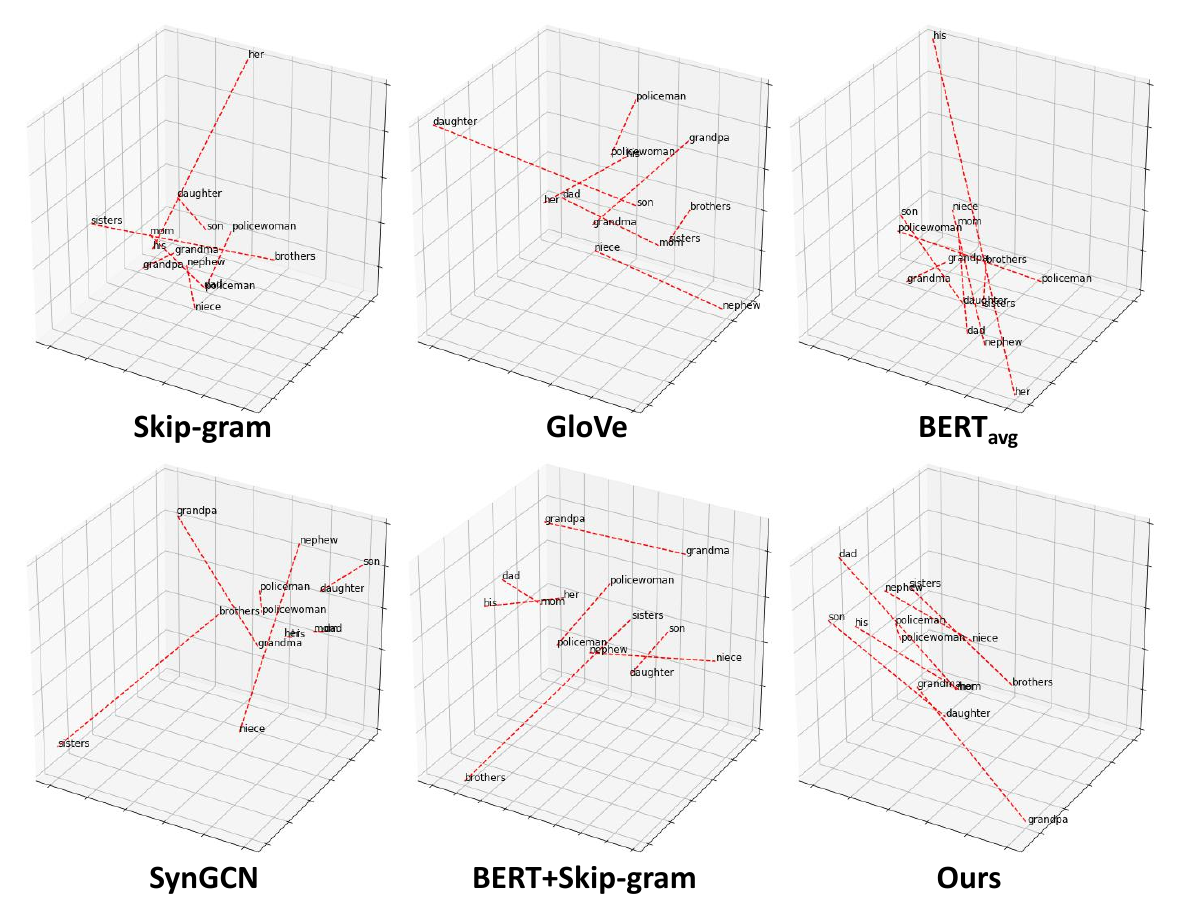}
	\caption{Visualization on word pairs of gender relationship.}
	\label{fig_4}
\end{figure}

\subsection{Quantitative Comparison}
\textbf{Word Similarity and Analogy}. Table \ref{table_1} shows the experimental results of intrinsic tasks. Overall, the models that integrate contextualized information into static embeddings (\textit{Context-to-Vec}) perform better than other types (\textit{Contextualized / Static}). Our results outperform baselines across the board. To be fair, the backbone of our model here is BERT as that in the main baseline (\textit{BERT+Skip-gram}) ~\citep{wang2021improving}.

Within the \textit{Context-to-Vec} category, our models perform best on all word similarity datasets. Our base model without post-processing obtains an average absolute improvement of about +23.8\%(+13.2) and related improvement of +4.4\%(+2.9) compared with the main baseline. The performance is further enhanced using post-processing with a +25.6\%(+14.2) absolute increase, and a +5.8\%(+3.8) relative increase compared with the main baseline, and a +1.4\%(+1.0) relative increase compared with our base model (w/o post-processing). It is worth mentioning that the main baseline does not perform better than BERT$_{avg}$ in \textit{Contextualized} group on the RG65 dataset, but our model does make up for their regrets, which indicates that our model is better at understanding contextual correlates of synonymy.

For the word analogy task, our performances are basically equal to the baselines. Overall, we gain the best score (+0.5) on the Google dataset but without a significant improvement. Although we do not gain the best score across all baselines on the SemEval dataset, our model performs better than the main baseline.

For different datasets, especially in word similarity tasks, the improvement of our preliminary model on WS353, SimiLex, RG65 (+4.1, +5.5, and +5.7, respectively) is significantly better than other datasets. For example, the improvement of the main baseline on the WS353R (relatedness) subset and the WS353 set is far greater than that on the WS353S (similarity) subset. While our model bridges their gaps in the WS353 set and also ensures that the performance of WS353S and WS353R is further improved slightly.

\textbf{Word Concept Categorization}. Word concept categorization is another important intrinsic evaluation metric. We use 4 commonly used datasets as shown in Table \ref{table_2}. Overall, our model without post-processing outperforms the baselines by a large margin, giving the best performance and obtaining an average performance gain of +5.2\%(+5.1) compared to the main baseline. In particular, the largest increases are observed on the ESSLI(N) (+7.5), ESSLI(V) (+3.8). And with post-processing, our model can obtain better improvements (+3.3 vs. +5.1). The experimental results show the advantage of integrating contextualized and word co-occurrence information, which can excel in grouping nominal concepts into natural categories.

\textbf{Extrinsic Tasks}. Extrinsic tasks reflect the effectiveness of embedded information through downstream tasks. We conduct extrinsic evaluation from chunking, NER, and POS tagging tasks as shown in Table \ref{table_3}. We select comparison representatives from the \textit{Static} group, the \textit{Contextualized} group, and the \textit{Context-to-Vec} group, respectively. Although the improvement is not significant compared with the intrinsic evaluations, it can be seen that our performances are better than the baselines, which can prove the superiority of our model. The primary baseline \textit{BERT+Skip-gram} obtains the second-best average score, but does not excel in the chunking task. In contrast, our model not only outperforms all baselines moderately on average, but also performs best in every individual task.

\subsection{Ablation and Analysis}
\textbf{Post-processing Schemes}. From Table \ref{table_1}, we can initially find that the post-processing method has a positive impact. To further quantitatively analyze, we compare more related methods as shown in Table \ref{table_4}. In this ablation experiment, the comparison baseline is our trained original word vectors (w/o retrofitting), and the other comparison methods include the singularity decomposition-based method~\citep{artetxe2018uncovering}, and the synonym-based constraint method~\citep{faruqui2014retrofitting}. From the results, we can see that other post-processing schemes can improve the word vectors to some extent, but do not perform better in all datasets. However, our proposed post-processing scheme performs the best across the board here, which shows that converting the distance distribution into a probability distribution is more effective.


\textbf{Nearest Neighbors}. To further understand the results, we show the nearest neighbors of the words "\textit{light}" and "\textit{while}" based on the cosine similarity, as shown in Table \ref{table_6}. For the noun "\textit{light}", other methods generate more noisy and irrelevant words, especially static embeddings. In contrast, the \textit{Context-to-Vec} approaches (Ours \& BERT+Skip-gram) can capture the key meaning and generate cleaner results, which are semantically directly related to "\textit{light}" literally. For the word "\textit{while}", the static approaches tend to co-occur with the word "\textit{while}", while \textit{Context-to-Vec} approaches return conjunctions with more similar meaning to "\textit{while}", such as "\textit{whilst}", "\textit{whereas}" and "\textit{although}", which demonstrates the advantage of using contextualization to resolve lexical ambiguity.

\textbf{Word Pairs Visualization}. Fig.\ref{fig_4} shows the 3D visualization of the gender-related word pairs based on t-SNE~\citep{van2008visualizing}. These word pairs differ only by gender, e.g., "\textit{nephew} vs. \textit{niece}" and "\textit{policeman} vs. \textit{policewoman}". From the topology of the visualized vectors, the spatial connectivity of the word pairs in Skip-gram and GloVe is rather inconsistent, which means that static word vectors are less capable of capturing gender analogies. In contrast, for vectors based on contextualized embeddings, such as BERT$_{avg}$, SynGCN, BERT+Skip-gram, and our model, the outputs are more consistent. In particular, our outputs are highly consistent in these instances, which illustrates the ability of our model to capture relational analogies better than baselines and the importance of contextualized information based on semantic knowledge.

\section{Conclusion}
We considered improving word embeddings by integrating more contextual information from existing pre-trained models into the Skip-gram framework. In addition, based on inherent properties of static embeddings, we proposed a graph-based post-retrofitting method by employing priori synonym knowledge and a weighted distribution probability. The experimental results show the superiority of our proposed methods, which gives the best results on a range of intrinsic and extrinsic tasks compared to baselines. In future work, we will consider prior knowledge directly during training to avoid a multi-stage process.

\section*{Acknowledgements}
This work is supported in part by the Science and Technology Innovation 2030 - Major Project (No. 2021ZD0150100) and National Natural Science Foundation of China (No. U21A20427). We thank all the anonymous reviewers for their helpful comments and suggestions.

\bibliography{anthology,custom}
\bibliographystyle{acl_natbib}

\appendix



\end{document}